\documentclass{article}
\usepackage{spconf,amsmath,graphicx}

\usepackage{booktabs}

\usepackage[utf8]{inputenc} 
\usepackage[T1]{fontenc}    
\usepackage{hyperref}       
\usepackage{url}            
\usepackage{booktabs}       
\usepackage{amsfonts}       
\usepackage{nicefrac}       
\usepackage{microtype}      
\usepackage{multirow}
\usepackage[table]{xcolor}
\usepackage{amssymb}
\usepackage{pifont}
\newcommand{\xmark}{\ding{55}}

\usepackage{diagbox}

\usepackage{bbm}
\usepackage[utf8]{inputenc} 
\usepackage[T1]{fontenc}    
\usepackage{hyperref}       
\usepackage{url}            
\usepackage{booktabs}       
\usepackage{nicefrac}       
\usepackage{microtype}      
\usepackage{color}
\usepackage{colortbl}
\usepackage[outercaption]{sidecap}
\usepackage{caption}
\definecolor{Gray}{gray}{0.90}
\definecolor{white}{rgb}{1.0, 1.0, 1.0}
\definecolor{LightCyan}{RGB}{247, 223, 231}
\newcolumntype{a}{>{\columncolor{LightCyan}}c}
\definecolor{Gray}{gray}{0.90}
\usepackage{tikz}
\usetikzlibrary{shapes.misc,shadows}
\usepackage{booktabs}
\usepackage{graphicx}
\usepackage{wrapfig}
\usepackage{rotating}
\usepackage{amsmath}
\usepackage{mathtools}
\usepackage{amssymb}
\usepackage{algpseudocode}
\usepackage{algorithm}
\usepackage{graphicx}
\usepackage{subfigure}

\title{Multi-Attribute Vision Transformers are \\ Efficient and Robust Learners}
\name{Hanan Gani, Nada Saadi, Noor Hussein, Karthik Nandakumar \thanks{All correspondences to hanan.ghani@mbzuai.ac.ae}}
\address{Mohamed Bin Zayed University of Artificial Intelligence (MBZUAI)}
\begin{document}
%
\maketitle
\begin{abstract}
Since their inception, Vision Transformers (ViTs) have emerged as a compelling alternative to Convolutional Neural Networks (CNNs) across a wide spectrum of tasks. ViTs exhibit notable characteristics, including global attention, resilience against occlusions, and adaptability to distribution shifts. One underexplored aspect of ViTs is their potential for multi-attribute learning, referring to their ability to simultaneously grasp multiple attribute-related tasks. In this paper, we delve into the multi-attribute learning capability of ViTs, presenting a straightforward yet effective strategy for training various attributes through a single ViT network as distinct tasks. We assess the resilience of multi-attribute ViTs against adversarial attacks and compare their performance against ViTs designed for single attributes. 
Moreover, we further evaluate the robustness of multi-attribute ViTs against a recent transformer based attack called Patch-Fool.
Our empirical findings on the CelebA dataset provide validation for our assertions.
\end{abstract}
\section{Introduction}
\label{sec:intro}

Multi-Task learning (MTL) or Multi-Attribute Learning (MAL) is a subset of machine learning in which multiple tasks or attributes are simultaneously learned by a shared model, which allows learning common ideas between a collection of related tasks. These shared representations increase data efficiency and can potentially yield faster learning speed for related or downstream tasks. In facial data, multi-attribute learning can be used to recognize and analyze various attributes of a face, such as age, gender, facial expressions, etc. The algorithm can then learn to identify the relationships between these attributes and the visual features of a face. During testing, the algorithm can use this information to make predictions about the attributes of a new facial image. 

 In case of multi-attribute learning, for a given sample, many of its attributes can be correlated. For example, if
we observe that a person has blond hair and heavy makeup, the probability of that person being attractive is high. Another example is that the attributes of beard and woman are almost impossible to appear on a person at
the same time. Modeling complex inter-attribute associations is an important
challenge for multi-attribute learning. To address this challenge, most existing
approaches adopt a multi-task learning framework, which formulates
multi-attribute recognition as a multi-label classification task and simultaneously
learns multiple binary classifiers.

Almost all of the previously proposed approaches in MAL are based on Convolutional Neural Networks (CNNs), which leaves quite a  room  for  exploring  this  problem  from  the  perspective of other vision architectures such as Vision Transformers. Since their inception, Vision Transformers (ViTs) \cite{dosovitskiy2020image} have emerged as an effective alternative to traditional Convolutional Nerual Networks (CNNs). ViT typically splits the image into a grid of non-overlapping patches before passing them to a linear projection layer to adjust the token dimensionality. These tokens are then processed by a series of feed-forward and multi-headed self-attention layers. Due to their ability to capture global structure through self-attention, ViTs have found extensive applications in many tasks. One specific advantage of the global self attention inside ViTs is the robustness against adversarial perturbations \cite{goodfellow2014explaining} which has been recently explored in the works of \cite{shao2021adversarial}, \cite{fu2022patch} and \cite{joshi2021adversarial}.

The research into adversarial attacks \cite{goodfellow2014explaining} on Vision Transformers (ViTs) has gained prominence with the escalating utilization of deep learning models in security-sensitive applications. In this paper, we take a focused approach to delve into the multi-attribute learning aspect of Vision Transformers, considering the evolving landscape shaped by the paradigm shift towards Transformers. Additionally, our investigation extends to the realm of adversarial attacks on these multi-attribute learning models. It is noteworthy that the exploration of adversarial attacks specifically on multi-attribute Vision Transformers represents a novel and unexplored domain. To the best of our knowledge, there is a lack of prior work in this particular area.


In this paper, we present two key contributions. We leverage the adaptability of Vision Transformers (ViTs) to accommodate variable number of tokens and propose a novel architectural design capable of handling multiple attribute tasks without compromising individual attribute performances. Specifically, we encode the task attributes as additional learnable tokens inside the ViT similar to the concept of prompts introduced in \cite{zhou2022learning}. However, in contrast to \cite{zhou2022learning}, our approach ensures the propagation of the attribute task tokens throughout the model alongside the patch tokens and are utilized to derive outputs for each attribute task. Subsequently, an in-depth analysis is undertaken to compare the robustness of multi-attribute ViTs with single-attribute ViTs, evaluating their resilience against well-established adversarial attacks. Additionally, we assess the robustness of multi-attribute ViTs against a recent patch-based attack tailored for transformers. Throughout these evaluations, empirical evidence is provided to underscore the robustness of our proposed multi-attribute ViT.

In summary, our contributions are as follows:
\begin{itemize}
    \item We introduce a novel architectural framework, termed MAL-ViT, designed specifically for Vision Transformers (ViTs). This proposed approach encodes attribute tasks as learnable tokens, facilitating the exchange of valuable information among various attributes. 
 \item Our proposed MAL-ViT is robust against various adversarial attacks. 
 \item We perform extensive analysis demonstrating the robustness of MAL-ViT compared to ViTs tarined on single attribute.

\end{itemize}

\section{Related Work}
\label{sec:format}
\textbf{Multi-Attribute Learning. }
 Multi-attribute learning has attracted increasing interest due to its broad applications \cite{cao2018partially,ak2018learning}. It involves many different
visual tasks \cite{hand2016attributes, liu2016deepfashion} according to the object of interest. Many works focus on
domain-specific network architectures. Cao et al. \cite{cao2018partially} proposed a partially shared
multi-task convolutional neural network (PS-MCNN) for face attribute recognition. The PS-MCNN consists of four task-specific networks and one shared
network to learn shared and task-specific representations. Zhang et al. \cite{9156843} proposed Two-Stream Networks for clothing classification and attribute recognition. Since some attributes are located in the local area of the image, many
methods \cite{guo2019visual, sarafianos2018deep} resort to the attention mechanism. Guo et al. \cite{guo2019visual} presented
a two-branch network and constrained the consistency between two attention
heatmaps. A multi-scale visual attention and aggregation method was introduced in \cite{sarafianos2018deep}, which extracted visual attention masks with only attribute-level
supervision. Tang et al. \cite{tang2019improving} proposed a flexible attribute localization module
to learn attribute-specific regional features. Some other methods \cite{kalayeh2017improving, liu2016deepfashion} further
attempt to use additional domain-specific guidance. Semantic segmentation was
employed in \cite{kalayeh2017improving} to guide the attention of the attribute prediction. Liu et al. \cite{liu2016deepfashion}
learned clothing attributes with additional landmark labels. Different from these approaches, we use a Vision transformer (ViT) model to learn attributes of facial images by modeling them as tokens inside ViT.

\noindent
\textbf{Adversarial Attacks:}
Szegedy et al. \cite{szegedy2013intriguing} first demonstrated the possibility of
fooling neural networks into making different predictions
for test images that are visually indistinguishable. Goodfellow et al. \cite{goodfellow2014explaining} introduced the one-step
Fast Gradient Sign Method (FGSM) prediction attack which
was followed by more effective iterative attacks Kurakin et al. \cite{kurakin2018adversarial}. Madri et al. \cite{madry2018towards} introduced Projected Gradient Descent (PGD) where the perturbations are iteratively applied to the input in the direction that maximizes the model's loss function. More recently, \cite{croce2020reliable} introduced AutoAttack which s a strong white-box adversarial attack that uses an ensemble of several diverse adversarial attack methods. In this paper, we utilize these existing attacks to validate the robustness of multi-attribute Vision Transformers.

\noindent
\textbf{Adversarial attacks on Vision Transformers: }A recent study on adversarial attacks on vision transformers, "Adversarial Token Attacks on Vision Transformers" \cite{joshi2021adversarial}, concluded that transformers are vulnerable to token attacks, even with low token budgets. Additionally, the tokens tend to compensate for patch perturbations smaller than their size and ResNets and MLP-Mixer outperform Transformers in token attacks. Another study, "Patch-Fool" \cite{fu2022patch}, aimed to investigate the robustness of ViTs against adversarial perturbations. The approach taken in Patch-Fool involved determining the most influential patch using attention scores, then replacing it with an adversarial patch added to the adversarial objective to cause the vision transformer to misclassify the image.
Given these interesting approaches, our proposed methodology aims to further enhance the understanding of ViTs' robustness. 

\section{Methodology}
\subsection{Preliminaries}
In this section, we discuss some of the preliminaries required to understand our metholdology and analysis.

\noindent{\textbf{Vision Transformer: }} Since their inception, Vision Transformers have proven to be the state-of-the-art neural network architectures outperforming the CNN's \cite{alexnet} on many vision tasks. The standard Transformer receives as input a 1D
sequence of token embeddings. To handle 2D images, the original input image $x \in \mathbb{R}^{H\times W\times C}$ is reshaped into a
sequence of flattened 2D patches $x_p \in \mathbb{R} ^
{N^2\times(P.C)}$, where (\textit{H}, \textit{W}) is the resolution of the original
image, C is the number of channels, (P, P) is the resolution of each image patch, and $N = HW/P^2$
is the resulting number of patches, which also serves as the effective input sequence length for the Transformer. The transformer uses constant latent vector size through all of its layers. The Transformer encoder (Vaswani et al., 2017) consists of alternating layers of multiheaded self-attention and MLP block. Layernorm (LN) is applied before
every block, and residual connections after every block.
The MLP contains two layers with a GELU non-linearity.

\noindent{\textbf{CLS Token: }} This is an extra learnable token added to the patch embeddings and is processed with the patch embeddings inside the ViT. It serves as representation of an entire image, which can be used for classification. Usually, the final classification output is taken from the CLS token.

\noindent{\textbf{Multi-Headed Self-Attention: }}The key concept behind self attention is that it allows the network to learn how best to route  information between the tokens. A standard attention mechanism extracts query $q$, key $k$ and value $v$ from the input $x$ using learnable weight matrices. The general idea is to take query vectors and match them against a series of key vectors that describe values  that we want to know something about. The similarity of a given query to the keys determines how much information from each value to retrieve for that particular query. The final attention equation is given as,

\begin{equation}
    Attention(q,k,v) = Softmax\big(\frac{qk^T}{\sqrt{d_k}}).v \\
\end{equation}
where $d_k$ is the dimension of the token embedding.

\subsection{Multi-attribute Learning in Vision Transformers}
In this section, we discuss our multi-attribute learning approach in ViTs. We start by discussing the concept of attribute tokens in section \ref{task_tokens} and then move to multi-attribute classification approach in section \ref{MAL-cls} 

\subsubsection{Attribute Tokens}
\label{task_tokens}
The transformer architecture is flexible to incorporating additional learnable embeddings apart from input embeddings. This characteristics of transformers has been greatly exploited to induce some important properties inside the ViT such as ability to capture shape, knowledge distillation etc. Generally in classification tasks, a learnable classification token called CLS token is added to the patch tokens. The CLS token captures the overall representation of the image as a whole and is commonly chosen to propagate features to the final classification layer. Similar to the CLS tokens, we introduce attribute tokens which capture the attribute information and model relationships between the attributes by interacting with each other and exchanging useful features inside multi-headed self-attention layers. We hypothesize that the introduction of attribute tokens in the ViTs helps in learning multiple attributes simultaneously without any further computational overhead. We verify this strategy empirically from the perspective of multi-task classification.

\begin{figure}[h!]
\centering
\includegraphics[width=8.2cm]{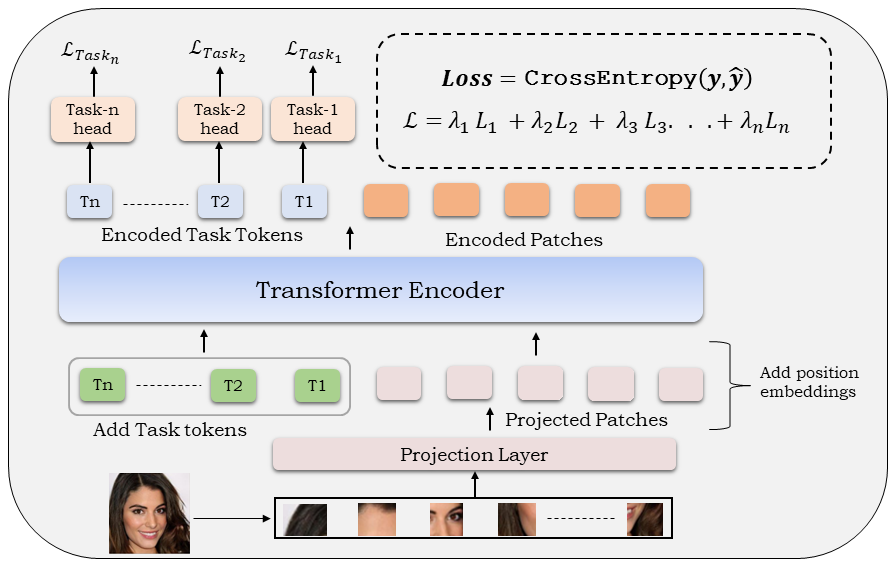}
  \caption{\small Our Multi-Attribute ViT Framework: We introduce additional learnable attribute (task) tokens corresponding to each attribute and propagate them jointly with patch tokens inside the ViT. We take the output corresponding to each attribute from its respective token. 
  }
\label{fig: main_diagram}
\end{figure}

\subsubsection{Multi-Attribute Classification}
\label{MAL-cls}
As mentioned above, we introduce the concept of attribute tokens to efficiently peeform multi-attribute learning in ViTs. In case of multi-attribute classification problem, we formulate the problem as n-binary classification tasks denoted as  $\mathcal{T}$.
\begin{equation}
    \centering
    \displaystyle \mathcal{T} = \{ \mathcal{T}_{1}, \mathcal{T}_{2}, \mathcal{T}_{3}, . . . , \mathcal{T}_{n} \}
\end{equation}
We introduce an attribute token for each classification task. As shown in Figure \ref{fig: main_diagram} a typical ViT breaks the input $x$ into small patches which are passed through a projection layer to obtain the corresponding patch embeddings.  At this stage, attribute tokens pertaining to each task in MAL framework are added to the input patches. The attribute tokens along with the patch embeddings are processed inside the transformer encoder, where they interact with each other inside the self-attention layers and transfer useful features among themselves with the help of attention mechanism. The final classification output for each task is taken from its corresponding attribute token. Let ${f}_{T}$ represent the features of the attribute tokens from the final transformer layer and $\mathcal{H}_{T}$ denote the final classification heads corresponding to each task. Then the output from each task head is represented as,
\begin{equation}
    \centering
    \displaystyle \hat{y}_{T} =\mathcal{H}_{T}({f}_{T})\,; \quad T = 1, \dots, n
\end{equation}
We use the commonly employed cross-entropy loss as the objective function to be minimized. The loss function for any task in the task set $\mathcal{T}$ is given as,
\begin{equation}
    \centering
    \displaystyle Loss = {\tt BinaryCrossEntropy} (y, \hat{y})     
\end{equation}
where $y$ and $\hat{y}$ are the true and predicted labels with respect to the given task.

The losses corresponding to each task are summed and back-propagated through the shared ViT backbone. The overall loss is given as, 
\begin{equation}
        \displaystyle \mathcal{L}_{total} =  \lambda_{1} \mathcal{L}_{1} +  \lambda_{2} \mathcal{L}_{2} +  \lambda_{3} \mathcal{L}_{3} + . . .  +  \lambda_{n} \mathcal{L}_{n}
\end{equation}
where $\mathcal{L}_{1}, \mathcal{L}_{2} , . . . , \mathcal{L}_{n}$ are the losses corresponding to Tasks $\mathcal{T}_{1}, \mathcal{T}_{2} , . . . , \mathcal{T}_{n}$  respectively and $\lambda_{1}, \lambda_{1}, . . . , \lambda_{n}$ are the learnable weights pertaining to each task loss. We validate the performance of our multi-attribute ViT network in Tables \ref{tab: celeba}.

\section{Experiments}

\subsection{Datasets}
We validate our approach and analysis on CelebA dataset, the details of which are shown below.
\\
\textbf{Celeb-A}  \cite{liu2015faceattributes} dataset, it is a large-scale dataset of celebrity faces, created for the purpose of training machine-learning models for facial recognition and other computer vision tasks. It contains over 200,000 images of more than 10,000 celebrities, annotated with facial landmarks and attributes such as age, gender, and facial expression. We select 9 attributes from the CelebA dataset which are given as: [5 O Clock Shadow, Black Hair, Blond Hair, Brown Hair,
Goatee, Mustache, No Beard, Rosy Cheeks, Wearing Hat].

\subsection{Evaluation Metrics}
\label{eval_strategy}
We measure the Multi-Attribute learning capabilities of ViT’s
under following settings.

\noindent \textit{MAL-ViT (ours) vs Single Attribute ViT (SAL-ViT):} We train all the
attribute tasks through a MAL-ViT and compare the performance of each task with its corresponding single Attribute ViT baseline. \\
\noindent \textit{MAL-ViT (ours) vs Multi-Attribute CNN’s:} We calculate the
percent increment / decrement in the performance of each
task when trained jointly with a CNN network Vs. when trained
jointly via MAL-ViT.

For adversarial robustness, we compare the \textit{robust accuracy} of \textit{multi-attribute ViT (MAL-ViT)} with the \textit{single-attribute ViT (SAL-ViT)}. Since our tasks are based on classification, we use classification accuracy and balanced accuracy as evaluation
metrics. 

\subsection{Training and Implementation details} 
We train all our models with Pytorch \cite{NEURIPS2019_bdbca288} using Pytorch lightning module.  We use auto LR scheduler with Adam optimizer \cite{kingma2014adam} with a batch size of 128 and set maximum epochs as 100. Additionally we use patience value as 10 epochs so that training stops as soon as the saturation occurs. We use ViT-Tiny architecture with patch size of 16 as the base architecture for ViT based networks. For CNN counterpart, we use ResNet-18 which roughly has same number of parameters as ViT-Tiny. For adversarial attacks, we use \textit{torchattacks} library. We keep steps=10 for the PGD attacks for both single-attribute and multi-attribute ViTs. For the hardware, we train our models on a single Nvidia A100 40GB GPU.

\section{Results}

\subsection{Multi-Attribute Learning Capability}
\label{Multi-Attribute learning capability}

\begin{figure}
\centering
\includegraphics[width=8cm]{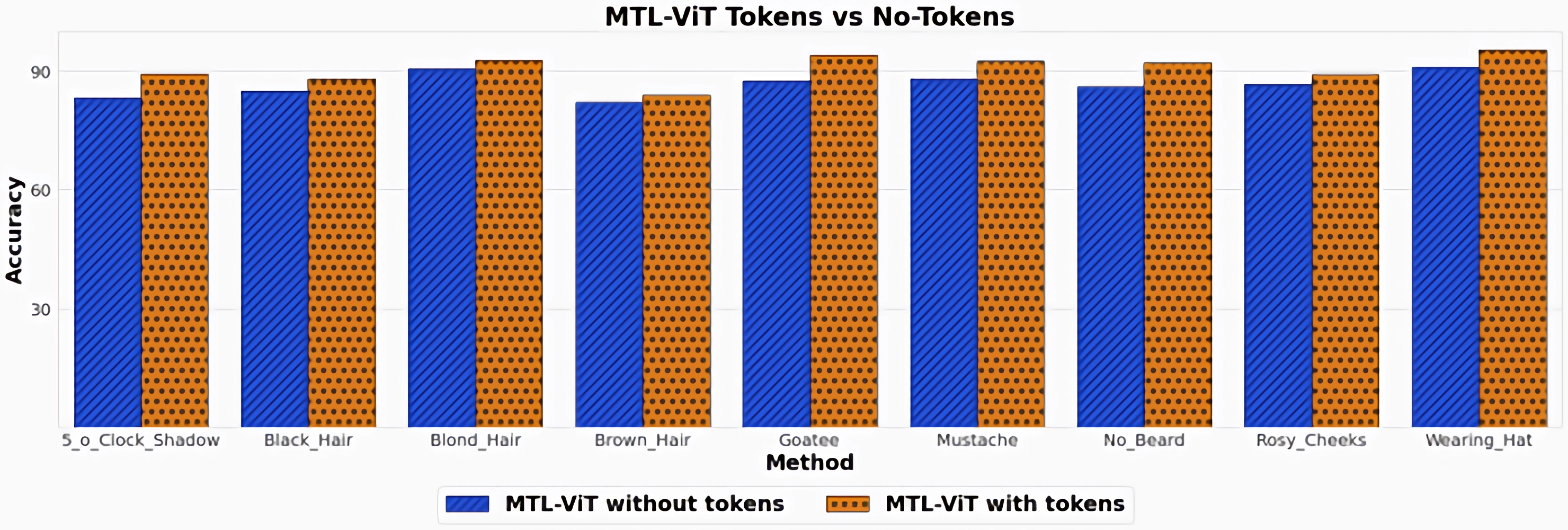}
  \caption{\small \textit{Effectiveness of task tokens}. Our method MAL-ViT with task tokens (shown in orange in the plot) outperforms the one without tokens. Best viewed in zoom.}
\label{fig: tasktokens-effectiveness}
\end{figure}

In this Table \ref{tab: celeba}, we compare our multi-attribute approach (Fig \ref{fig: main_diagram}) against the Convolutional neural networks along with single-attribute ViTs. First, we show that the relative increment in the balanced accuracy from single-task vs multi-task is higher in case of our multi-attribute ViT compared to CNNs. This shows that our MAL-ViT is efficient in learning multiple tasks together. We further validate the effectiveness of our multi-attribute ViT by comparing it against single-attribute ViT in second row. We showcase the efficacy of our approach via a notable increase in accuracy for each individual task. This shows that our MAL-ViT with attribute tokens effectively models the relationships between the attributes using minimal computational budget compared to training all the attributes separately. To further show the effectiveness of our proposed learnable attribute tokens, we train the MAL-ViT with and without the attribute tokens. Fig. \ref{fig: tasktokens-effectiveness} shows the our MAL-ViT with attribute tokens outperforms the one withou tokens when jointly trained on all the attributes.

\begin{table}
  \begin{minipage}{0.6\linewidth}
   \centering 
  \setlength{\tabcolsep}{11pt}
  \scalebox{0.38}[0.38]{
  \begin{tabular}{c|cccccccccc} \toprule
  \rowcolor{Gray}
    \rotatebox{90}{\textbf{Method}} & \rotatebox{90}{\textbf{5oClock}} & \rotatebox{90}{\textbf{Black Hair}} & \rotatebox{90}{\textbf{Blond Hair}} & \rotatebox{90}{\textbf{Brown Hair}} &  \rotatebox{90}{\textbf{Goatee}} & \rotatebox{90}{\textbf{Mustache}} &  \rotatebox{90}{\textbf{No Beard}} & \rotatebox{90}{\textbf{Rosy Cheeks}} & \rotatebox{90}{\textbf{Wearing Hat}} \\
    \midrule
   MAL-CNN vs 1-task CNN & +0.941\% & +1.22\% & -3.92\% & \textbf{+2.98\%} & -1.51\% & -12.08\% & \textbf{+1.53\%} & -4.15\% & -1.02\% \\
   \\
   MAL-ViT \textbf{(Ours)} vs 1-task ViT  & \textbf{+2.12\%} & \textbf{+2.21\%} & \textbf{+0.248\%} & +1.3\% & \textbf{+3.57\%} & \textbf{+1.17\%} & +1.5\% & \textbf{+0.86\%} & \textbf{+1.15\%} \\
  \bottomrule
  \end{tabular}}
  \end{minipage}
 
  \caption{\small \textit{Relative increment in the scores when training all attributes jointly vs training individually.} Our MAL-ViT outperforms MAL-CNN, showing that our MAL-ViT is an efficient multi-attribute learner.}
\label{tab: celeba}
\end{table}

\subsection{MAL-ViT vs SAL-ViT: Adversarial Robustness}
\label{SAL vs MAL: Adversarial Robustness}
\noindent

\begin{table*}[h!]
  \begin{minipage}{1.0\linewidth}
   \begin{center} 
  \setlength{\tabcolsep}{11pt}
  \scalebox{0.7}[0.7]{
  \begin{tabular}{c|c|ccccccccc|c} \toprule
  \rowcolor{Gray}
    \textbf{Method} & \textbf{Attack} & \rotatebox{90}{\textbf{5oClock}} & \rotatebox{90}{\textbf{Black Hair}} & \rotatebox{90}{\textbf{Blond Hair}} & \rotatebox{90}{\textbf{Brown Hair}} &  \rotatebox{90}{\textbf{Goatee}} & \rotatebox{90}{\textbf{Mustache}} &  \rotatebox{90}{\textbf{No Beard}} & \rotatebox{90}{\textbf{Rosy Cheeks}} & \rotatebox{90}{\textbf{Wearing Hat}} & \rotatebox{90}{\textbf{Avg Robust Acc.}}\\
    \midrule
     & \xmark & 88.5 & 86.8 & 92.8 & 83.4 & 93.3 & 91.9 & 91.8 & 88.9 &  95.6 &90.3\\
     \multirow{6}{*}{SAL-VIT} & FGSM & 1.8 & 17.0 & 0.0 & 4.0 & 5.0 & 2.0 & 1.0 & 5.0 & 9.0 & 4.9\\
     &  BIM ($\epsilon = 0.03$) & 0.07 & 10.3 & 0.01 & 0.5 & 1.5 & 0.01 & 0.07 & 0.2 & 0.09 & 1.41\\
     & PGD (steps = 10) & 0.0 & 11.0 & 0.0 & 1.0 & 1.0 & 1.0 & 0.0 & 18.0 & 0.0 &3.5\\ 
    & UAP (epochs=5) & 52.3 &49.9 &50.0 &49.9 &49.6 &49.9 &49.9 &49.9 & 49.9 &50.1\\
    & UAP (epochs=10) &52.3 &49.9 &50.0 &49.9 &50.9 &49.6 &49.9 &49.9 &49.9 & 50.2\\
    & Patch-Fool &31.1 &31.7 &41.7 &36.0 &32.2 &38.0 &28.7 &46.1 &60.5 &38.4\\
     \midrule
     & \xmark  & 90.3 & 88.7 & 93.1 & 84.48 & 96.6 & 93.3 & 90.2 & 96.4 & 92.0 & \textbf{91.6}\\ 
     \multirow{6}{*}{MAL-ViT \textbf{(Ours)}} & FGSM & 20.3 & 49.6 & 30.6 & 29.3 & 28.9 & 26.7 & 26.4 & 48.9 & 30.9 & \textbf{32.4}\\
     & BIM ($\epsilon = 0.03$)& 20.4 & 49.6 & 29.2 & 28.0 & 28.6 & 26.4 & 26.1 & 48.7 & 29.8 &\textbf{31.8}\\
    & PGD (steps = 10) & 20.3 & 49.6 & 30.6 & 29.2 & 28.9 & 26.7 & 26.4 & 48.9 & 31.0 &\textbf{32.4}\\
    & UAP (epochs = 5) & 59.9 & 64.8 & 54.6 & 49.9 & 56.2 & 53.0 & 56.3 & 51.6 & 67.4  &\textbf{57.0}\\
    & UAP (epochs = 10) & 59.3 & 64.7 & 54.5 & 49.9 & 57.3 & 55.5 & 55.8 & 51.0 & 67.7 &\textbf{57.3}\\
     & Patch-Fool & 49.4 & 73.2 & 72.8 & 66.9 & 53.6 & 54.2 & 58.0 & 57.5 & 65.0 &\textbf{61.1}\\
  \bottomrule
  \end{tabular}}
  \end{center}
  \end{minipage}
  \vspace{0.2em}
  \caption{Comparison of robustness of single-attribute ViT (SAL-ViT) vs multi-attribute ViT (MAL-ViT) using existing adversarial attacks and patch based attack under different settings. All scores are reported on balanced accuracy (\%). As seen, our MAL-ViT is more robust to adversarial attacks.}
  \vspace{-0.5em}
\label{tab:basicattacks}
\end{table*}

\begin{figure}[h!]
  \centering
  {\includegraphics[width=0.7\linewidth]{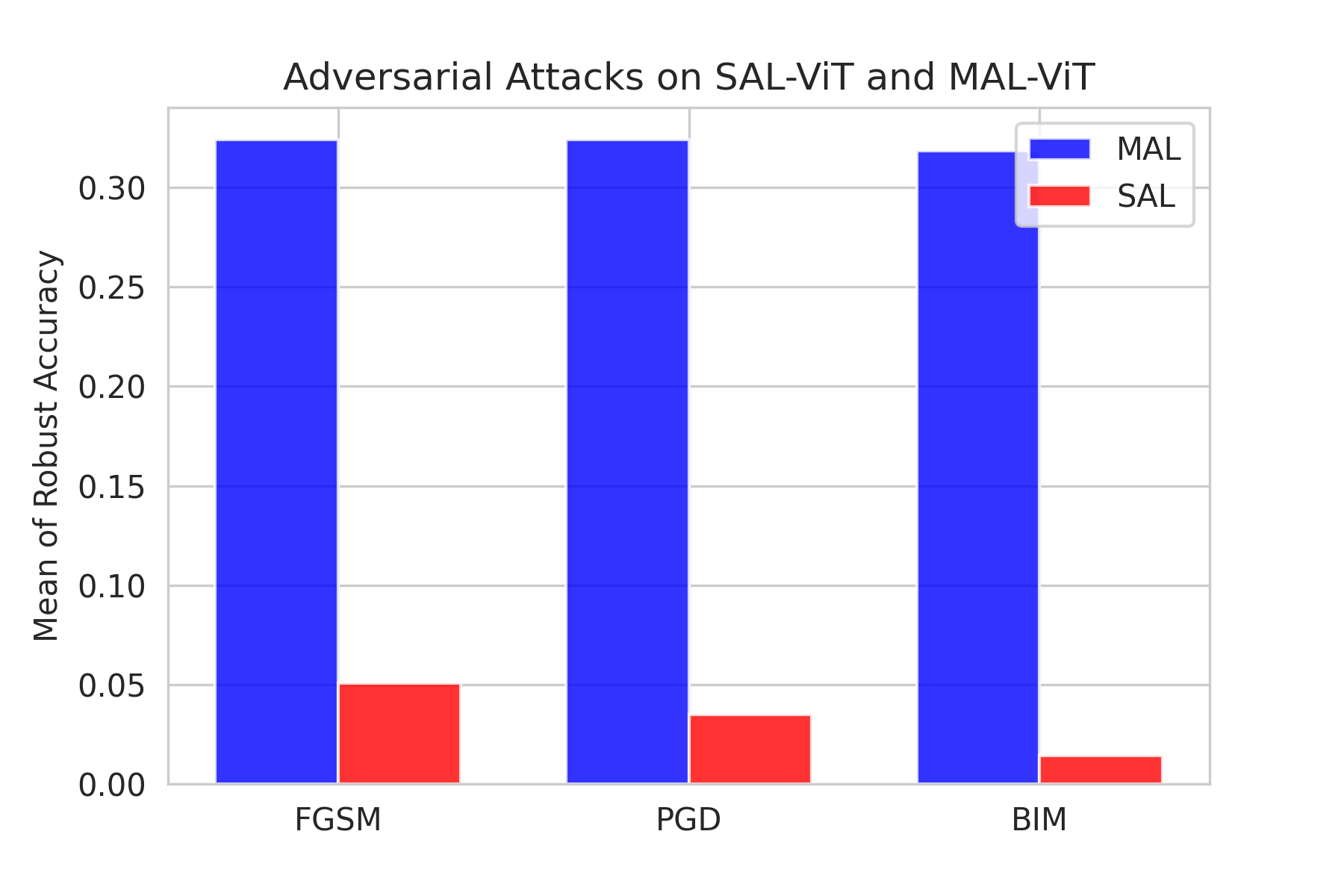}}
  \caption{Each bar represent the mean of the robust accuracy of MAL-ViT (blue) and MAL-ViT (red), when attacking the model with FGSM, PGD and BIM}
  \label{fig:mean-adv-barplot}
\end{figure}

\textbf{Robustness against common adversarial attacks.} 
In Table \ref{tab:basicattacks}, our study entails a series of diverse attacks applied to SAL-ViT and MAL-ViT, specifically Fast Gradient Sign Method (FGSM), Projected Gradient Descent (PGD) attack and Basic Iterative attack (BIM). Through our experimentation (Table \ref{tab:basicattacks}), we observe that in all the cases our MAL-ViT has a higher robust accuracy compared to SAL-ViT when exposed to these attacks. For instance, in case of PGD attack, for attributes such as '5oClock', 'Blond Hair', and 'No beard', the robustness score is zero in case of single-attribute ViT. However, in multi-attribute scenario, the robust accuracy is above 20\% for all the cases. Similar trend can be observed in FGSM and BIM attacks, where our multi-attribute ViT outperforms the single-attribute ViT in all the cases. Similarily, with BIM attack, we observe a similar trend when experimented with multiple epsilon ($\epsilon$) values. Fig \ref{fig:mean-adv-barplot} shows the performance comparison of MAL-ViT vs SAL-ViT in terms of mean robust accuracy over all the attributes.

\begin{figure}[h!]
  \centering
  {\includegraphics[width=0.7\linewidth]{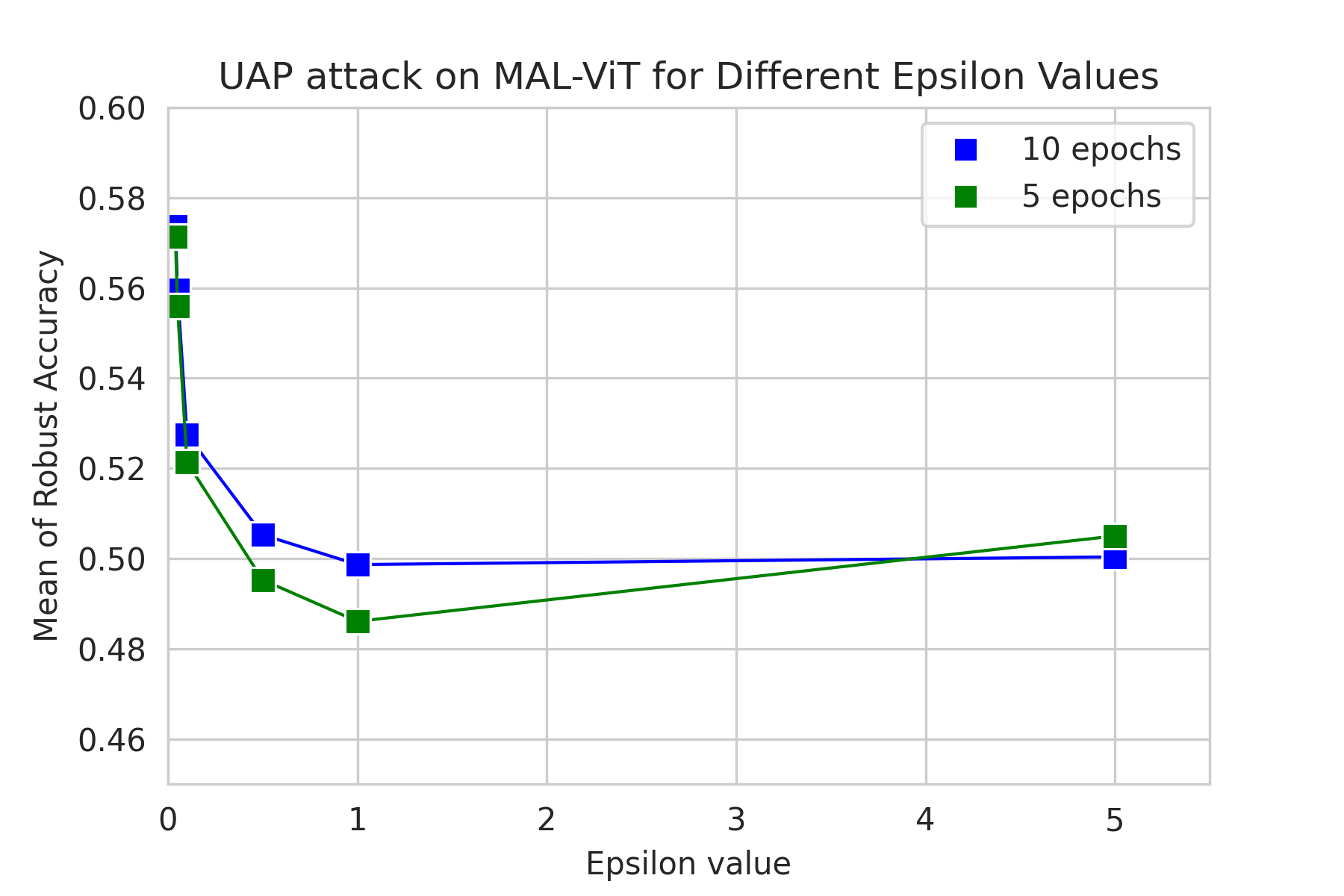}}
  \caption{MAL-ViT under UAP attack with different epsilon values}
  \label{fig:uap-barplot}
\end{figure}

We further utilize Universal Adversarial Perturbations (UAPs) to verify the robustness of our approach. As evident from Table \ref{tab:basicattacks}, both MAL-ViT and SAL-ViT models exhibit a degree of resilience against the UAP attack, with the robust accuracy hovering around 0.5 for the attributes of both models. However, it is noteworthy that despite the UAP being a more generalized attack, it proves less effective when applied to ViTs. Furthermore, our observations indicate that MAL-ViT consistently outperforms SAL-ViT in terms of robust accuracy across all attributes. This substantiates the claim that our model demonstrates superior robustness compared to SAL-ViT, particularly in the face of a more broadly applicable UAP attack. In Figure \ref{fig:uap-barplot}, we present a comparative analysis of the robust accuracy of MAL-ViT under various epsilon values. The UAP attack is most effective when $\epsilon = 1$ in experiments with both 5 and 10 epochs. Importantly, MAL-ViT exhibits robustness against the UAP attack for values of $\epsilon$ both below and above 1, showcasing a higher mean robust accuracy.

\begin{figure}[h!]
  \centering
  {\includegraphics[width=0.7\linewidth]{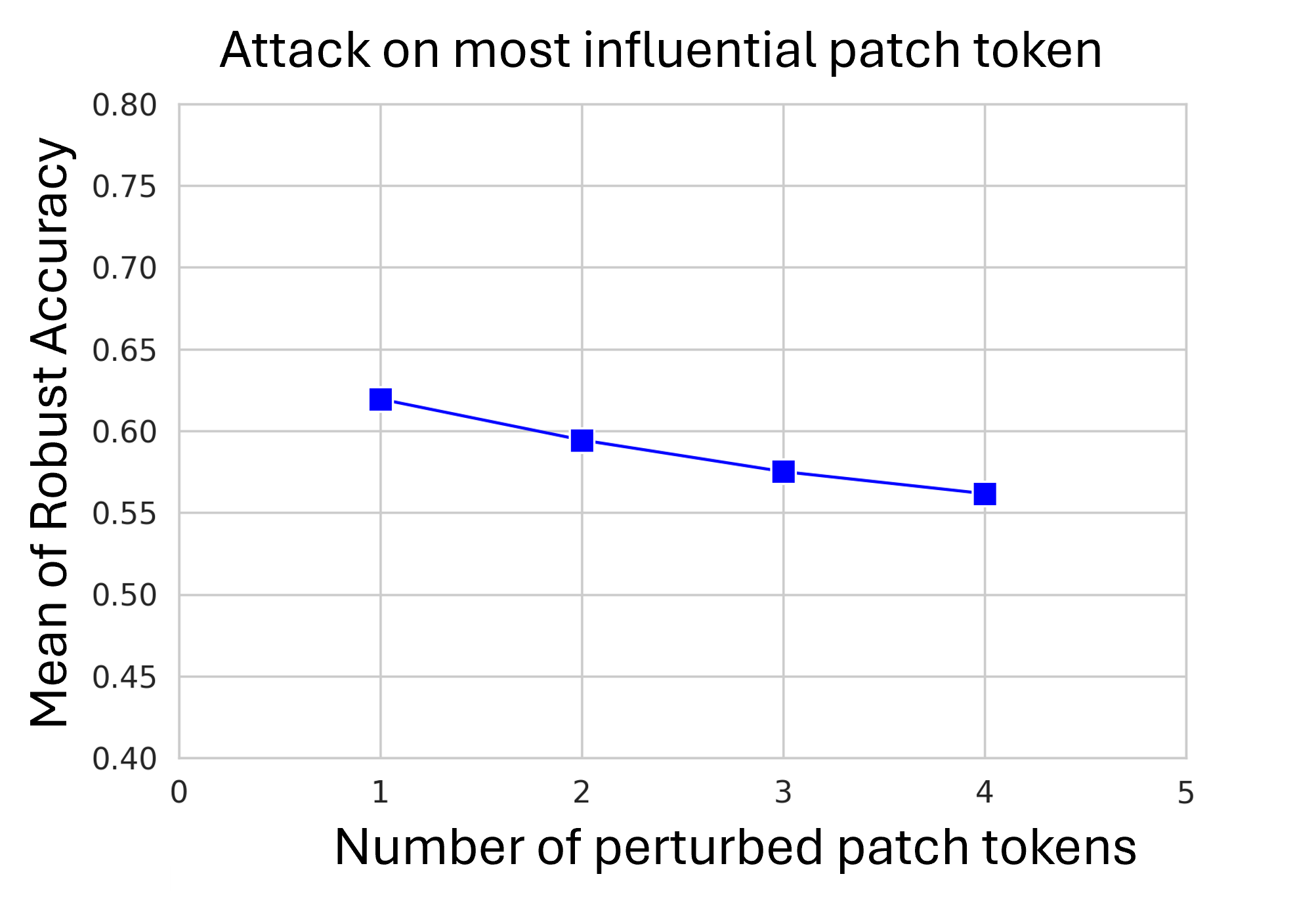}}
  \caption{Mean robust accuracy vs. number of perturbed patch tokens under Patch-Fool attack on MAL-ViT.}
  \label{fig:patchfool-lineplot}
\end{figure}

\noindent
\textbf{Robustness against patch attacks.}
Table \ref{tab:basicattacks} presents the outcomes of patch-fool attack on both SAL-ViT and MAL-ViT by perturbing only a single most influential patch based on attention score. The results indicate that MAL-ViT is more resilient to patch-based attacks, demonstrating higher robust accuracy in comparison to SAL-ViT. 
In Figure \ref{fig:patchfool-lineplot} We offer an analysis by progressively increasing the number of perturbed patches, which results in a decline in the mean balanced accuracy for all attributes. However, even with the increase in the number of perturbed patches, the accuracy of MAL-ViT does not decline as much as SAL-ViT.

\section{Discussion and future work}
We propose that our multi-attribute ViT (MAL-ViT) framework in addition to being efficient than single-attribute ViT (SAL-ViT), is robust to adversarial attacks. As a first step to verify our claim, we designed a setup comparing the adversarial robustness of MAL-ViT with SAL-ViT against common existing adversarial attacks such as FGSM, PGD, BIM and UAP, and patch based attacks such as Patch-Fool. We hypothesize that the interactions between the attributes inside the ViTs self-attention mechanism makes it robust against the existing white box attacks relative to the single-attribute model. Additionally, the incorporation of loss from different attributes enhances the overall robustness of MAL-ViT, establishing its resilience relative to its single-attribute counterparts. In the future, our plan is to expand the multi-attribute learning capability of MAL-ViT to other tasks, such as segmentation and detection. 

\section{Conclusion}

In this paper, we propose a novel design which demonstrates multi-attribute learning capability in ViTs by incorporating additional learnable tokens for each attribute task. The effectiveness of our Multi-attribute ViT (MAL-ViT) becomes evident as it outperforms CNN and single-attribute ViT (SAL-ViT) in concurrently learning multiple attributes. Subsequently, we scrutinize the robustness of MAL-ViT, evaluating its performance against SAL-ViT across various attacks and parameter settings. Our empirical findings consistently highlight that MAL-ViT displays superior robustness by showcasing enhanced performance under diverse attacks and parameter configurations.

\bibliography{refs}
\bibliographystyle{abbrv}

\end{document}